\documentclass[runningheads,final]{llncs}

\usepackage{graphicx,graphics}
\usepackage[usenames,dvipsnames]{color}
\usepackage{url}
\usepackage{amsmath,amsfonts,bbm,amssymb,textcomp,booktabs,threeparttable}
\usepackage{hhline}
\usepackage{multirow}
\usepackage[dvipsnames]{xcolor}
\usepackage[export]{adjustbox}
\usepackage[bookmarks=false,bookmarksopen=false,breaklinks=true,colorlinks=true, linkcolor=blue,anchorcolor=blue,citecolor=blue,filecolor=blue,menucolor=blue, urlcolor=blue]{hyperref}
\usepackage{stmaryrd,rotating}
\include{boldcommands}
\usepackage{tabularx}
\usepackage{booktabs}
\usepackage{eucal}

\newcommand{\tss}{\textsuperscript}

\begin{document}
\mainmatter 

\title{Synthesizing CT from Ultrashort Echo-Time MR Images via Convolutional Neural Networks}
\titlerunning{CT Synthesis from UTE via CNN}

\author{Snehashis Roy\textsuperscript{1} 
\thanks{\small{Support for this work included funding from the Department of Defense in the
Center for Neuroscience and Regenerative Medicine.}}
\and John A. Butman\textsuperscript{2} 
\and Dzung L. Pham\textsuperscript{1}
}
\authorrunning{S. Roy \textit{et. al.}}

\institute{\textsuperscript{1} 
Center for Neuroscience and Regenerative Medicine, Henry Jackson Foundation, USA\\
\textsuperscript{2} Radiology and Imaging Sciences, Clinical Center, National Institute of Health, USA\\
}

\maketitle

\begin{abstract}
With the increasing popularity of PET-MR scanners in clinical applications, synthesis of CT images 
from MR has been an important research topic. Accurate PET image reconstruction requires attenuation 
correction, which is based on the electron density of tissues and can be obtained from CT images. 
While CT measures electron density information for x-ray photons, MR images convery information 
about the magnetic properties of tissues.  Therefore, with the advent of PET-MR systems, the 
attenuation coefficients  need to be indirectly estimated from MR images. In this paper, we propose 
a fully convolutional neural network (CNN) based method to synthesize head CT from ultra-short 
echo-time (UTE) dual-echo MR images. Unlike traditional $T_1$-w images which do not have any bone 
signal, UTE images show some signal for bone, which makes it a good candidate for MR to CT 
synthesis. A notable advantage of our approach is that accurate results were 
achieved with a small training data set. Using an atlas of a single CT and dual-echo UTE pair, we 
train a deep neural network model to learn the transform of
MR intensities to CT using patches. We compared our CNN based model with a state-of-the-art 
registration based as well as a Bayesian model based CT synthesis method, and showed that the 
proposed CNN model outperforms both of them. We also compared the proposed model when only $T_1$-w 
images are available instead of UTE, and show that UTE images produce better synthesis 
than using just $T_1$-w images.

\end{abstract}

\section{Introduction}
\label{sec:intro}
Accurate PET (positron emission tomography) image reconstruction requires correction for the 
attenuation of $\gamma$ photons by tissue. The attenuation coefficients, called $\mu$-maps, can be 
estimated from CT images, which are x-ray derived estimates of electron densities in tissues. 
Therefore PET-CT scanners 
are well suited for accurate PET reconstruction. In recent years, PET-MR scanners have become more 
popular in clinical settings. This is because of the fact that unlike CT, MRI (magnetic resonance 
imaging) does not impart any radiation, and MR images have superior soft tissue contrast. However, 
an MR image voxel contains information about the magnetic properties of the tissues at that voxel, 
which has no direct relation to its electron density. Therefore synthesizing CT from MRI is an 
active area of research.

Several MR to CT synthesis methods for brain images have been proposed. Most of them can be 
categorized into two classes -- segmentation based and atlas based. CT image intensities
represent quantitative Hounsfeld Units (HU) and their standardized values are usually known for air, 
water, bone, and other brain tissues such as fat, muscle, grey matter (GM), white matter (WM), 
cerebro-spinal fluid (CSF) etc. Segmentation based methods 
\cite{martinez2009,berker2012} first segment a $T_1$-w MR image of the whole head into 
multiple classes, such as bone, air, GM, and WM. Then each of the segmented classes are replaced 
with the mean HU for that tissue class, or the intensity at a voxel is obtained from the 
distribution of HU for the tissue type of that voxel.

Most segmentation based approaches rely on accurate multi-class segmentation of $T_1$-w images. 
However, traditional $T_1$-w images do not produce any signal for bone. As bone has the highest 
average HU compared to other soft tissues, accurate segmentation of bone is crucial for accurate PET 
reconstruction. 
Atlas based methods \cite{hofmann2011} can overcome this limitation via registration. An atlas 
usually consists of an MR and a co-registered CT pair. For a new subject, multiple atlas MR images 
can be deformably registered to the subject MR; then the deformed atlas CT images are combined using 
voxel based label fusion \cite{burgos2014} to generate a synthetic subject CT. It has 
been shown that atlas based methods generally outperform segmentation based methods 
\cite{hofmann2011}, because they do not need accurate segmentation of tissue classes, which can be 
difficult because it becomes indistinguishable from background, tissues with short $T_1$, and 
tissues whose signal may be suppressed, such as CSF. 

One disadvantage of registration based methods is that a large number of atlases is needed for
accurate synthesis. For example, $40$ atlases were used in \cite{burgos2014}, leading to 
significantly high computational cost with such a large number of registrations. To alleviate this 
problem, atlas based patch matching methods have been proposed \cite{roy2014,malpica2016}. For a 
particular patch on a subject MR, relevant matching patches are found from atlas MR images. The 
atlases only need to be rigidly registered to the subject \cite{malpica2016}. The matching atlas MR 
patches can either be found from a neighborhood of that subject MR patch \cite{malpica2016}, or from 
any location within the head \cite{roy2014,roy2014spie}. Once the matching patches are found, their 
corresponding CT patches are averaged with weights based on the patch similarity to form a 
synthetic CT. The advantage of patch matching is that deformable registration is not needed, thereby 
decreasing the computational burden and increasing robustness to differences in the anatomical 
shape. These type of methods also require fewer atlases (e.g., $10$ in \cite{malpica2016} and $1$ in 
\cite{roy2014}).

Recently, convolutional neural networks (CNN), or deep learning \cite{hinton2015}, has been 
extensively used in many medical imaging applications, such as lesions and tumor segmentation
\cite{kamnitsas2017}, brain segmentation, image synthesis, and skull stripping. Unlike traditional 
machine learning algorithms, CNN models do not need hand-crafted features, and are therefore 
generalizable to a variety of problems. They can accommodate whole images or much larger patches 
(e.g., $17^3$ in \cite{kamnitsas2017}) compared to smaller sized patches used in most patch based 
methods (e.g., $3^3$ in \cite{malpica2016}), thereby introducing better neighborhood 
information. A CNN model based on U-nets \cite{ronnenberger2015} has been recently proposed to 
synthesize CT from $T_1$-w images \cite{han2017}. In this paper, we propose a synthesis method based 
on fully convolutional neural networks to generate CT images from dual-echo UTE images. We 
compare with two leading CT synthesis methods, one registration based \cite{burgos2014} and one 
patch based \cite{roy2014}, and show that our CNN model produces more accurate results compared to 
both of them. We also show  that better synthesis can be obtained using UTE images rather than
only $T_1$-w images.

\section{Data Description}
MR images were acquired on $7$ patients on a 3T Siemens Biograph mMR. The MR acquisition includes
$T_1$-w dual-echo UTE and MPRAGE images. The specifications of UTE images are as follows,
image size $192 \times 192 \times 192$, resolution $1.56$ mm\tss{3}, repetition time $TR=11.94$s, 
echo time $TE=70\mu$s and $2.46$ms, flip angle $10^\circ$. MPRAGE images were acquired 
with the following parameters, resolution $1.0$ mm\tss{3}, $TR=2.53$s, $TE=3.03$ms, flip angle 
$7^\circ$. CT images were acquired on a Biograph $128$ Siemens PET/CT scanner with a tube voltage of 
$120$ kVp, with dimensions of $512 \times 512 \times 149$, and resolution of $0.58 \times 0.58 
\times 1.5$ mm\tss{3}. MPRAGE and CT were rigidly registered \cite{avants2011} to the second UTE 
image. All MR images were corrected for intensity inhomogeneities by N4 \cite{tustison2010}. The 
necks were then removed from the MPRAGE images using FSL's \texttt{robustfov} \cite{jenkinson2012}. 
Finally, to create a mask of the whole head, background noise was removed from the MPRAGE using 
Otsu's threshold \cite{otsu1979}. UTE and CT images were masked by the headmask obtained from the 
corresponding MPRAGE. Note that the choice of MPRAGE to create the headmask is arbitrary, CT could 
also be used as well. The headmask was used for two purposes.
\begin{enumerate}
\item Training patches were obtained within the headmask, so that the center voxel of a patch 
contains either skull or brain.
\item Error metrics between synthetic CT and the original CT were computed only within the headmask.
\end{enumerate}

\begin{figure}[!bt]
\begin{center}
\includegraphics[width=1\textwidth]{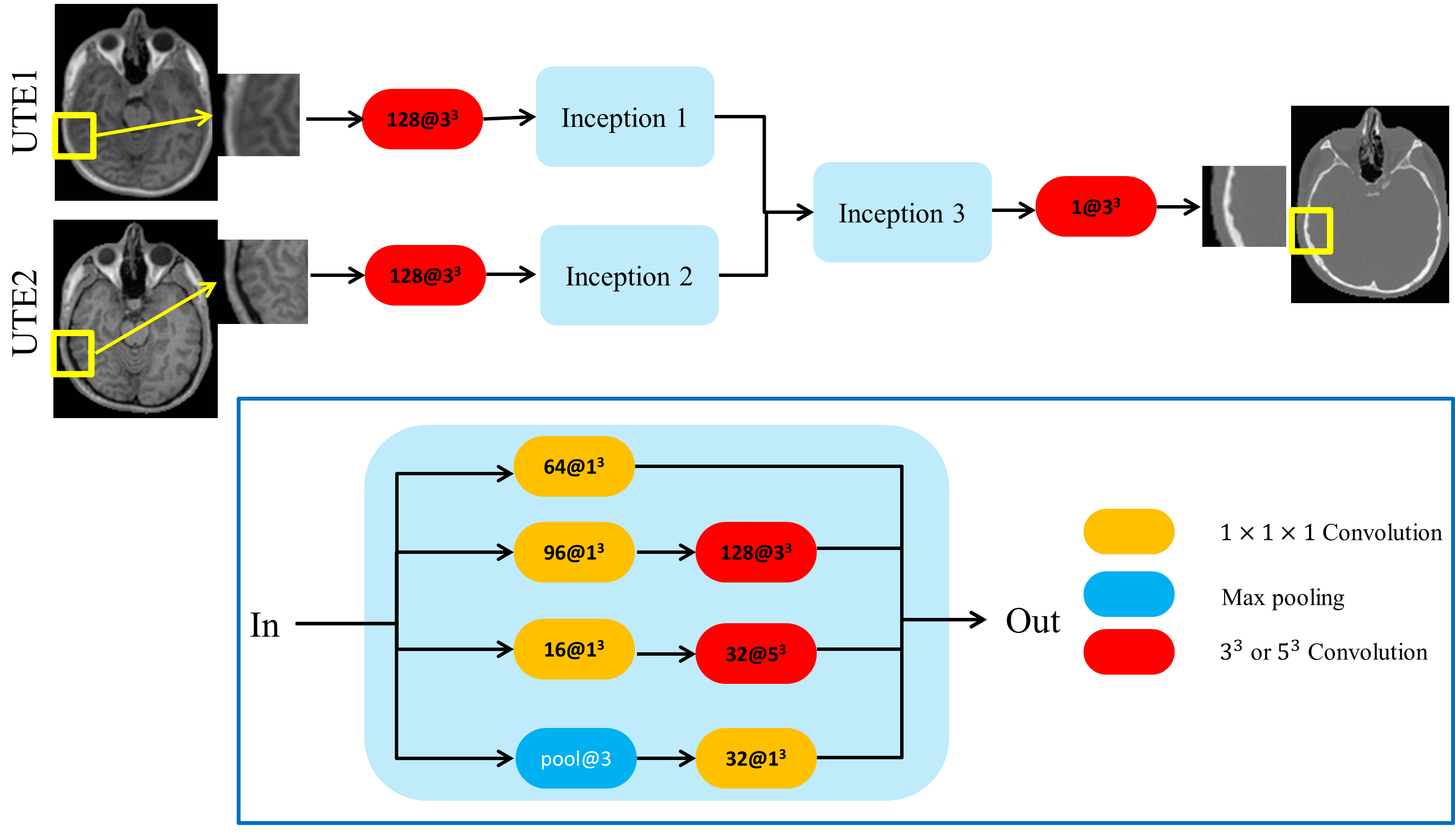} 
\end{center}
\caption{The figure shows the proposed CNN model incorporating the Google Inception block
\cite{szegedy2015}, shown in inset. During training, patches from each of the dual-echo UTE images 
are first independently processed through two Inception blocks. Then their outputs are concatenated 
and again processed through another Inception block. Finally, the mean squared errors between the
CT patch and the output of the third Inception block is minimized to train the parameters of the 
CNN. A convolution is written as $128@3^3$, indicating there are $128$
filters of size $3\times 3\times 3$. The pooling layer is defined as \texttt{pool@3}, indicating
maximum value within a $3\times 3\times 3$ region is used. Convolutions and pooling are done
with stride $1$. All convolutions are followed by ReLU, although for brevity, they are not shown 
here.
}
\label{fig:Fig1}
\end{figure}

\section{Method}
\label{sec:method}
We propose a deep CNN model to synthesize CT from UTE images. Although theoretically the model can 
be used with whole images, we used patches due to memory limitations. Many CNN architectures have 
previously been proposed. In this paper, we adopt Inception blocks \cite{szegedy2015}, that have 
been successfully used in many image classification and recognition problems in natural image 
processing  via GoogleNet. The rationale for using this architecture over U-net 
is discussed in Sec.~\ref{sec:discussion}. The proposed CNN architecture is shown in 
Fig.~\ref{fig:Fig1}. 

Convolutions and pooling are two basic building layers of any CNN model. Traditionally they 
are used in a linear manner, e.g. in text classification \cite{lecun1998}. The primary innovation
of the Inception module \cite{szegedy2015} was to use them in a parallel fashion. In an Inception 
module, there are two types of convolutions, one with traditional $n^3 (n>1)$ filter banks, and one 
with $1^3$ filter banks. It is noted that $1^3$ filters are downsampling the number of channels. The 
$1^3$ filters are used to separate initial number of channels ($128$) into multiple smaller sets 
($96$, $16$, and $64$). Then the spatial correlation is extracted via $n^3 (n>1)$ filters. The 
downsampling of channels and parallelization of layers reduce the total number of parameters to be 
estimated,which in turn introduces more non-linearity, thereby improving classification accuracy
\cite{chollet2016,szegedy2015}. Note that the proposed model is fully convolutional, as we did not 
use a fully connected layer.

During training, $25 \times 25\times 5$ patches around each voxel within the headmask are extracted 
from the UTE images with stride $1$. Then the patches from each UTE image are first convolved with 
$128$ filters of size $3\times 3\times 3$. Such a filter is denoted by $128@3^3$ in 
Fig.~\ref{fig:Fig1}. The outputs of the
filters are processed through separate Inception blocks. The outputs of these Inception blocks are 
then concatenated through their channel axis (which is same as the filter axis)
and processed through another Inception block and a $3^3$ filter. The coefficients 
of all the filters are computed by minimizing mean squared errors between the CT patch and the 
output of the model via stochastic gradient descent. Note that every convolution is followed by a 
ReLU (rectified linear unit), module, which is not shown in the figure. The pooling is performed by 
replacing each voxel of a feature map by the maximum of its $3\times 3\times 3$ neighbors.

\begin{figure}[!tbh]
\begin{center}
\tabcolsep 1pt
\begin{tabular}{ccccc}
& \texttt{UTE 1} & \texttt{UTE 2} & \texttt{MPRAGE} & \texttt{CT} \\
\rotatebox{90}{\hspace{3em}Subject \#1} &
\includegraphics[width=0.22\textwidth]{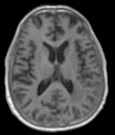}  &
\includegraphics[width=0.22\textwidth]{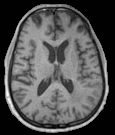}  &
\includegraphics[width=0.22\textwidth]{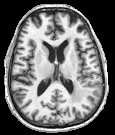}  &
\includegraphics[width=0.22\textwidth]{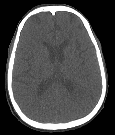}  \\
& \texttt{Fusion} & \texttt{GENESIS} & \texttt{CNN w/ MPRAGE} & \texttt{CNN w/ UTE} \\
\rotatebox{90}{\hspace{3em}Subject \#1} &
\includegraphics[width=0.22\textwidth]{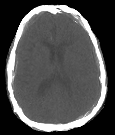}  &
\includegraphics[width=0.22\textwidth]{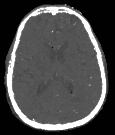}  &
\includegraphics[width=0.22\textwidth]{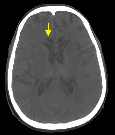}  &
\includegraphics[width=0.22\textwidth]{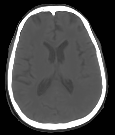}  \\

& \texttt{UTE 1} & \texttt{UTE 2} & \texttt{MPRAGE} & \texttt{CT} \\
\rotatebox{90}{\hspace{3em}Subject \#2} &
\includegraphics[width=0.22\textwidth]{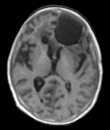}  &
\includegraphics[width=0.22\textwidth]{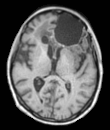}  &
\includegraphics[width=0.22\textwidth]{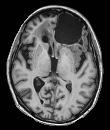}  &
\includegraphics[width=0.22\textwidth]{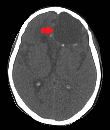}  \\
& \texttt{Fusion} & \texttt{GENESIS} & \texttt{CNN w/ MPRAGE} & \texttt{CNN w/ UTE} \\
\rotatebox{90}{\hspace{3em}Subject \#2} &
\includegraphics[width=0.22\textwidth]{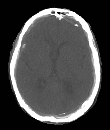}  &
\includegraphics[width=0.22\textwidth]{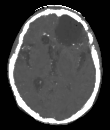}  &
\includegraphics[width=0.22\textwidth]{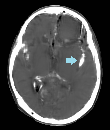}  &
\includegraphics[width=0.22\textwidth]{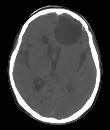}  
\end{tabular}
\end{center}
\caption{Top two rows show UTE, MPRAGE, original, and synthetic CT images of a healthy volunteer.
Bottom two rows show the same for a patient with a large lesion. Fusion \cite{burgos2014} shows
diffused bone in subject \#1, while the CNN with MPRAGE shows some artifacts near ventricles
(yellow arrow). Both GENESIS and the synthetic CT obtained with UTE can successfully reproduce the 
lesion (red arrow) for subject \#2, with CNN synthesis showing less noise.
}
\label{fig:Fig2}
\end{figure}

We used Adam \cite{kingma2015} as the 
optimizer to estimate the filter weights. Adam has been shown to produce much faster convergence
than comparable optimizers. While training, we used $75$\% of the total atlas patches as training 
set and $25$\% as the validation set. To obtain convergence, $25$ epochs were used. The filter
parameters were initialized by randomly choosing numbers from a zero-mean Gaussian distribution
with standard deviation of $0.001$. A batch size of $64$ was empirically chosen and found to 
produce sufficient convergence without requiring much GPU memory. The model was implemented in Caffe 
\cite{jia2014}. Anisotropic $25\times 25\times 5$ patches were used because larger size isotropic 
patches requires more GPU memory, while the patch size was empirically estimated. To compensate for 
the fact that patches are anisotropic, the atlas was reoriented in three 
different orientations -- axial, coronal, and sagittal. Training was performed separately for
each oriented atlas to generate three models, one for each orientation. Then for a new subject, the 
models were applied on the corresponding reoriented versions of the subject, and then averaged to 
generate a mean synthetic CT. Training on a TITAN X GPU with $12$GB memory takes about $6$ hours. 
Synthesizing a CT image from a new subject takes about $30$ seconds, where approximately $10$
seconds is needed to predict one orientation. Although the training is
performed using $25\times 25\times 5$ patches, the learnt models are able to predict a whole 2D 
slice of the image by applying the convolutions on every slice. Each of the three learnt models
were used to predict every 2D slice of the image in each of the three orientations. Then the $3$
predicted images were averaged to obtain the final synthetic CT.

\section{Results}

We compared our CNN based method to two algorithms, GENESIS \cite{roy2014} and intensity fusion
\cite{burgos2014}. GENESIS uses dual-echo UTE images and generates a synthetic CT based on another 
pair of UTE images as atlases. While GENESIS is a patch matching method which does not need any 
subject to atlas registration, the intensity fusion method (called ``Fusion") registers atlas 
$T_1$-w images to a subject $T_1$-w image, and combines the registered atlas CT images based on 
locally normalized correlation. In our implementation of Fusion, the second echo of an UTE image
pair was chosen as the subject image and was registered to the second echo UTE images of the 
atlases. The second echo was chosen for registration as its contrast closely 
matches the regular $T_1$-w contrast used in \cite{burgos2014}.
Similar to \cite{han2017} which proposed a CNN model only using $T_1$-w images, we also 
compared the proposed model with both channels as the MPRAGE.

One patient was arbitrarily chosen to be the ``atlas" for both GENESIS and the proposed CNN model
with both UTE and MPRAGE as inputs. The trained CNN models are applied to the other $6$ subjects. 
Since Fusion requires multiple atlas registrations, the validation is computed in a leave-one-out 
manner only for Fusion. GENESIS was also trained on the same atlas and evaluated on the remaining 
$6$.

Fig.~\ref{fig:Fig2} shows examples of two subjects, one healthy volunteer and one with a large
lesion in the left frontal cortex. For the healthy volunteer, all of the three methods perform 
similarly, while Fusion shows some diffused bone. It is because the deformable registrations can be
erroneous, especially in presence of skull. CNN with MPRAGE shows some artifacts near ventricles
(yellow arrow), while CNN with UTE images provide the closest representation to the original CT. For 
the subject with a brain lesion, Fusion can not successfully reproduce the lesion, as none of the 
atlases have any lesion in that region. CNN with MPRAGE shows artifacts where CSF is misrepresented 
as bone (blue arrow). This can be explained by the fact that both CSF and MPRAGE have low signal on 
MPRAGE. Synthetic CT from CNN with UTE shows the closest match to the CT, followed by GENESIS, which 
is noisier.

\begin{table}[!tb]
\caption{Quantitative comparison based on PSNR and linear correlation is shown for the competing
methods on $6$ subjects. Bold indicates largest value among the four synthetic CTs.
}
\begin{center}
\begin{tabular}{llcccccc}
\toprule[2pt]
& & \multicolumn{6}{c}{Subject \#}  \\
\cmidrule(lr){3-8}
\texttt{Metric} & \texttt{Method} & \texttt{1} & \texttt{2} & \texttt{3} & \texttt{4} & \texttt{5} 
& \texttt{6} \\
\toprule[1pt]
PSNR  & Fusion & 20.66 & 14.34 & 17.87 & 20.11 & 20.45 & 19.90 \\
      & GENESIS & 18.89 & 16.28 & 17.20 & 17.96 & 21.52 & 21.17 \\
      & CNN w/ MPRAGE & 22.35 & 16.46 & 16.00 & 22.06 & 21.91 & 21.32 \\
      & CNN w/ UTE & \textbf{23.40} & \textbf{18.76} & \textbf{19.78} & \textbf{23.49} & 
\textbf{23.54} & \textbf{22.54} \\
\toprule[1pt]
Correlation  & Fusion & 0.7377 & 0.6325 & 0.8097 & 0.7482 & 0.6807 & 0.6506 \\
      & GENESIS & 0.5852 & 0.6800 & 0.7747 & 0.6277 & 0.6875 & 0.7132 \\
      & CNN w/ MPRAGE & 0.7851 & 0.6995 & 0.6867 & 0.8007 & 0.7160 & 0.7137\\
      & CNN w/ UTE & \textbf{0.8384} & \textbf{0.8457} & \textbf{0.8820} & \textbf{0.8634} &
\textbf{0.8174} & \textbf{0.8017} \\
\bottomrule[2pt]
\end{tabular}
\end{center}
\label{tab:comparison}
\end{table}

To quantitatively compare the competing methods, we used PSNR and linear correlation coefficient
between the original CT and the synthetic CTs. PSNR is defined as a measure of mean squared error 
between original CT $\mathcal{A}$ and a synthetic CT $\mathcal{B}$ as, 
PSNR$=10\log_{10}(\frac{MAX_{\mathcal{A}}^2}{||\mathcal{A}-\mathcal{B}||^2})$,
where $MAX_\mathcal{A}$ denotes the maximum value of the image $\mathcal{A}$. Larger PSNR indicates
better matching between $\mathcal{A}$ and $\mathcal{B}$. Table~\ref{tab:comparison} shows the PSNR 
and correlation for Fusion, GENESIS, the proposed CNN model with only MPRAGE and with dual-echo UTE 
images. The proposed model with UTE images produces the largest PSNR and correlation compared
to both GENESIS and Fusion, as well as CNN with MPRAGE. A Wilcoxon signed rank test showed a
p-value of $0.0312$ comparing CNN with UTE with the other three for both PSNR and correlation,
indicating significant improvement in CT synthesis. Note that we used only $6$ 
atlases for our implementation of Fusion, although the original paper \cite{burgos2014} recommended 
$40$ atlases. Better performance would likely have been achieved with additional 
atlases.  Nevertheless, the proposed model outperforms it with only one atlas.

\section{Discussion}
\label{sec:discussion}
We have proposed a deep convolutional neural network model to synthesize CT from dual-echo UTE
images. The advantage of a CNN model is that 
prediction of a new image takes less than a minute. This efficiency is especially useful in clinical 
scenarios when using PET-MR systems, where PET attenuation correction is immediately 
needed after MR acquisition. Another advantage of the CNN model is that no atlas registration is 
required. Although adding multiple atlases can increase the training time linearly, the prediction
time ($\sim 30$ seconds) is not affected by the number of atlases. This is significant in comparison 
with patch based \cite{roy2014,malpica2016} and registration based approaches \cite{burgos2014} 
($\sim 1$ hr), where adding more atlases increases the prediction time linearly.

The primary limitation of the proposed, or in general, any CNN model is that it requires 
large amount of training data because the number of free parameters to estimate is usually large. In 
our case, by using only $3$ Inception modules, the total number of free parameters are approximately 
$29,000$. We used all patches inside the headmask which was about $500,000$ for the $1.56$ mm\tss{3} 
UTE images. By adding more Inception modules, as done in GoogleNet \cite{szegedy2015}, the number of 
free parameters grow exponentially, which needs more training data.  An important 
advantage of the proposed model over the U-net in \cite{han2017} is that only a single UTE image 
pair was used as atlas. Since we used patches instead of 2D slices \cite{ronnenberger2015,han2017} 
for training, the number of training samples is not limited by the number of slices in an atlas. 
One atlas with $256$ slices was used to generate $500,000$ training samples, which was 
sufficient to produce better results than competing methods. In clinical applications, it can be
difficult to obtain UTE and high resolution CT images for many subjects. Therefore using patches 
instead of slices give exponentially more training samples.

The patch size ($25\times 25\times 5$ ) is an important parameter of the model which was chosen 
empirically to make best practical use of the available GPU memory. Although CNN models do not need 
hand-crafted features, it was observed that using bigger patches usually increases accuracy. 
However, there lies a trade-off between patch size and available memory. Future work includes
optimization of patch size and number of atlases, as well as exploring further CNN architectures.

\bibliographystyle{splncs03}
\small{
\bibliography{refs}
}

\end{document}